\documentclass[12pt]{article}
\usepackage[usenames,dvipsnames]{color}

\usepackage{graphicx}
\usepackage{graphics}
\usepackage{amsmath,,amssymb}
\usepackage{algorithmic}
\usepackage{algorithm}
\usepackage{caption}
\usepackage{subcaption}
\usepackage{pstricks, pst-node}

\newtheorem{theorem}{Theorem}

\newtheorem{lemma}{Lemma}

\def\bE{\mathbb{E}}
\def\bP{\mathbb{P}}
\def\bR{\mathbb{R}}

\def\Fcal{\mathcal F}

\def\Hcal{\mathcal H}

\def\Scal{\mathcal S}

\usepackage{accents}
\newlength{\dhatheight}

\def\blist#1#2#3
{    \begin{list}{}{
     \setlength{\parsep}{0pt}
     \setlength{\leftmargin}{#1 pt}
     \setlength{\listparindent}{0pt}
     \setlength{\itemindent}{\listparindent}
     \setlength{\labelsep}{#3 pt}
     \setlength{\labelwidth}{\leftmargin}
     \addtolength{\labelwidth}{-\labelsep}
     \addtolength{\labelwidth}{\itemindent}
     \setlength{\rightmargin}{#2 pt}   }   }

\def\elist{\end{list}}

\def\bmu{\mbox{\boldmath $\mu$}}
\def\bzeta{\mbox{\boldmath $\zeta$}}
\def\zero{\mathbf 0}

\begin{document}
\title{Rate-Optimal Detection of Very Short Signal Segments}

\date{(\today)}

\author{T. Tony Cai$^\ast$ ~and~ Ming Yuan$^\dag$\\
University of Pennsylvania and University of Wisconsin-Madison}

\footnotetext[1]{
Department of Statistics, The Wharton School, University of Pennsylvania, Philadelphia, PA 19104. 
The research of Tony Cai was supported in part by NSF Grant DMS-1208982 and NIH Grant R01 CA127334.}
\footnotetext[2]{
Department of Statistics, University of Wisconsin-Madison, Madison, WI 53706. The research of Ming Yuan was supported in part by NSF Career Award DMS-1321692 and FRG Grant DMS-1265202.}

\maketitle

\begin{abstract}
Motivated by a range of applications in engineering and genomics, we consider in this paper detection of very short signal segments in three settings: signals with known shape, arbitrary signals, and smooth signals. Optimal rates of detection are established for the three cases and rate-optimal detectors are constructed.  The detectors are  easily implementable and are based on scanning with linear and quadratic statistics. Our analysis reveals both similarities and differences in the strategy and fundamental difficulty of detection among these three settings.
\end{abstract}

\noindent{\bf Keywords:} high-dimensional inference, optimal rate, scan statistics, signal detection, signal segments.

\noindent{\bf AMS 2000 Subject Classification:} Primary: 60G35; Secondary: 62G20.

\newpage

\section{Introduction}
Detection of very short signal segments arise in a wide range of applications in many fields including engineering, genomics, and material science. For example, copy number variations (CNVs)  play a significant role in the genetics of complex disease. Therefore the detection of CNVs due to duplication and deletion of a segment of DNA sequences is an important problem in genomics. In contrast to single-nucleotide polymorphisms which affects only one single nucleotide base, each CNV corresponds to a short segment of the genome, typically around 1000 nucleotide bases, that has been altered (see, e.g., Stankiewicz and Lupski, 2010). Although the length of these CNVs  is much smaller than that of the whole genome, recognizing and accounting for such segment structure are critical in effective detection of CNVs (see, e.g., Jeng, Cai and Li, 2010). Similar problems and phenomena also naturally arise in many other engineering and biological applications where the signal can be a moving target in video surveillance (see, e.g., NRC, 1995), geometric objects in computer vision (see, e.g., Arias-Castro, Donoho and Huo, 2005),  fissures in materials (Mahadevan and Casasent 2001), peaks associated with transcription factor binding sites in ChIP-Seq data (see, e.g., Schwartzman, et al., 2013), or change in the light curve of a star due to transiting planets (see, e.g., Fabrycky et al., 2012).

Motivated by the CNV analysis in genomics, detection of short, sparse, and piecewise constant segments have been well studied. See, for example, Arias-Castro, Donoho and Huo (2005), Zhang and Siegmund (2007), Jeng, Cai and Li (2010), Cai, Jeng and Li (2012), and the references therein. For a range of other applications mentioned above, the signal segments are not piecewise constant and the methods developed for detecting constant segments cannot be applied. In this paper, we consider detection of general sparse signal segments in three settings: signals with a known shape, arbitrary signals, and smooth signals. 

\subsection{Detection of Signal Segments}

The detection problem can be characterized by a signal-plus-noise model where observations $X_1,\ldots, X_n$ follow
$$ 
X_i=\mu_i+\epsilon_i, \qquad i=1,2\ldots, n,
$$
and $\epsilon_i\sim N(0,\sigma^2)$ is independent measurement error. In the absence of signal, 
$$
H_0: \mu_1=\mu_2=\ldots=\mu_n=0;
$$
while if signals are present, there is at least one segment $S=(a,b]$ for some $0\le a< b\le n$ not known a priori such that
\begin{equation}
\label{eq:signal}
H_1: \mu_i=f((i-a)/d)\qquad {\rm if\ } a< i\le b
\end{equation}
for an unknown function $f\in \Fcal$ where $\Fcal$ is a family of functions defined over $[0,1]$ and $d=b-a$. We are  interested in the problems of detection: When are such signal segments detectable? And how can they be effectively detected? Motivated by the applications mentioned earlier, we focus on very short signal segments in that $d$ diverges with $n$ such that $d<n^\xi$ for some $\xi<1$.

The problem of signal detection can be cast as testing the null hypothesis $H_0$ against the alternative $H_1$. We say that a signal is detectable if there exists a consistent test, that is, there exists a test whose type I and II errors both converge to zero. We investigate specifically three different settings -- when the shape of the signal is known in advance; when the signal is completely unknown; and when the signal is only known to be smooth. Optimal rates of detection are established for the three cases and easily implementable, rate-optimal detectors are constructed. Our analysis reveals profound similarities and differences in both the strategy and fundamental difficulty of detection among these three settings.

\subsection{Summary of Results}
In particular, it is shown that, in the first two settings, the detectability of a signal is determined jointly by its amplitude $A=(\int f^2)^{1/2}$ and the length of its duration $d:=b-a$. Specifically, if the shape of a signal is known in advance, the optimal rate of detection is 
\[
A\sim d^{-1/2}\log^{1/2} n
\]
in the sense that there exist constants $\overline{c}\ge \underline{c}>0$ and a detector such that any signal with amplitude $A>\overline{c}d^{-1/2}\log^{1/2} n$ can be identified by this detector; and conversely, if $A<\underline{c}d^{-1/2}\log^{1/2} n$, then the signal cannot be reliably identified by any detector, or as we shall formally describe later, there is no consistent test for $H_0$ against $H_1$. In contrast, without any information about the signal a priori, the optimal rate of detection is
$$
A\sim \left\{\begin{array}{ll}d^{-1/2}\log^{1/2} n & {\rm if\ } d=O(\log n)\\ d^{-1/4}\log^{1/4} n & {\rm if\ }d\gg \log n\end{array}\right.,
$$ 
which exhibits a phase transition at $d\asymp\log n$. For shorter signals, the optimal rate of detection of signal, knowing or not knowing its shape, is $A\sim d^{-1/2}\log^{1/2} n$; and surprisingly, there is no loss in terms of detection rate for not knowing the shape of a signal a priori. On the other hand, for longer signals, detection of signals of known shape is possible if $A\ge \overline{c}d^{-1/2}\log^{1/2} n$ for some constant $\overline{c}>0$; whereas detection of signals without any prior information is only possible if their amplitude is at least of the order $d^{-1/4}\log^{1/4} n$, indicating that the information on the shape of the signal can be extremely beneficial to its detection. Moreover, in both scenarios, the optimal rate of detection is attainable by scanning through all possible signal segments -- for each putative segment, an appropriate statistic is computed to summarize its likelihood of containing a signal; and the presence of a signal is claimed if and only if the maximum of all these statistics exceeds a given threshold. The choice of the statistic used in the scan, however, differs between the two cases. For signals of known shape, a linear statistic is used; whereas for unknown signals, a quadratic statistic is to be used.

Although in many applications, it may not be realistic to expect prior knowledge of its shape in advance, the signal may not be entirely unknown either. It is often reasonable to assume that the signal is smooth (see, e.g., Schwartzman, Garvrilov and Adler, 2011). It turns out that such qualitative information about the signal could help significantly to improve our ability of detecting the signal. More specifically, assume that the signal $f$ in (\ref{eq:signal}) is $\alpha$ times differentiable in that it belongs to the H\"older space of order $\alpha  \; (>0)$. Then the optimal rate of detection of the signal is
$$
A\sim \left\{\begin{array}{ll}d^{-1/2}\log^{1/2} n & {\rm if\ }d=O((\log n)^{2\alpha+1})\\ d^{-{2\alpha\over 4\alpha+1}}(\log n)^{\alpha\over 4\alpha+1}& {\rm if\ } d\gg (\log n)^{2\alpha+1}\end{array}\right.,
$$
when $\alpha\ge 1/4$; and
$$
A\sim \left\{\begin{array}{ll}d^{-1/2}\log^{1/2} n & {\rm if\ }d=O((\log n)^{2\alpha+1})\\ d^{-{2\alpha\over 4\alpha+1}}(\log n)^{\alpha\over 4\alpha+1}& {\rm if\ } d\gg (\log n)^{2\alpha+1} {\rm \ and\ } d=O((\log n)^{1/(1-4\alpha)})\\ d^{-1/4}\log^{1/4} n& {\rm if\ } d\gg (\log n)^{1/(1-4\alpha)}\end{array}\right.,
$$
when $\alpha<1/4$. In both cases, the loss of detection rate for not knowing a signal's shape only occurs when its length $d$ is of order greater than $(\log n)^{2\alpha+1}$. Another interesting observation is that when $\alpha\ge 1/4$, smoothness is always beneficial for longer signals; whereas when $\alpha<1/4$, the effect of smoothness vanishes if $d\gg (\log n)^{1/(1-4\alpha)}$, in which case detecting smooth signals is as difficult as detecting an arbitrary signal. In other words, for signals coming from a  H\"older space with $\alpha<1/4$, the knowledge of smoothness is only useful for signals of intermediate length. In addition, it is shown that the optimal rate of detection is attained through scanning all possible signal segments, with a hybrid of the linear and quadratic statistics that takes advantage of both statistics.

It is interesting to compare our results on detecting smooth signals with those from the work of Ingster (1993) or Ingster and Suslina (2003) who studied, to put in our context, optimal detection of a smooth signal at a known location and showed that the optimal rate of detection is $A^2\sim d^{-4\alpha/(4\alpha+1)}$ regardless of the length $d$ and degree $\alpha$ of smoothness of a signal. It is evident from our results that the effect of not knowing the location of a signal is very complex and leads to phase transition in the effect of both $d$ and $\alpha$. In particular, it is interesting to note that when $\alpha\ge 1/4$ and a signal is long, the effect of not knowing its location actually decreases with the degree of smoothness.

\begin{figure}[htbp]
\centerline{\includegraphics[width=0.7\textwidth,angle=270]{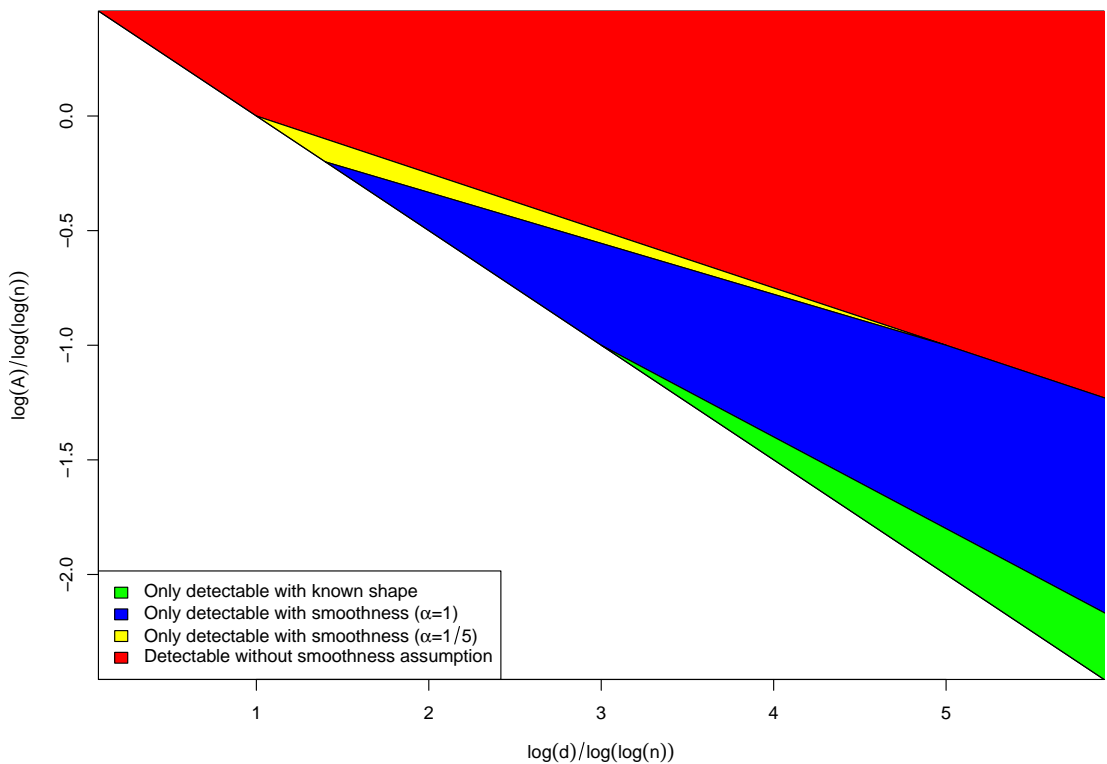}}
\vspace{-.2in}
\caption{\small Detection boundary for signals of arbitrary shape, known shape, or different degrees of smoothness. All shaded region corresponds to signals that are detectable if their shape is known in advance. Red, Yellow and Blue shaded regions are detectable if signals are known to be once differentiable ($\alpha=1$). Yellow and Red shaded regions are detectable if the signal is known to be from the H\"older space with $\alpha=1/5$. Red shaded region is detectable if no information of the signal is known in advance.}
\label{fig:boundary}
\end{figure}

These optimal rates of detection with different types of information are illustrated in Figure \ref{fig:boundary}.
In a $\log A/\log \log n$ versus $\log d/\log \log n$ plot, the optimal detection boundary for signals of known shape is the area above the diagonal, that is, all shaded areas in Figure \ref{fig:boundary}. In contrast, for arbitrary signals, the detection is only possible for signals that lie in the red quadrilateral in Figure \ref{fig:boundary}. In contrast, if we know a priori that the signal is from the H\"older space with $\alpha=1/5$, then the area of detection is the pentagon shaded in either red or yellow. Similarly, if the signal is from the H\"older space with $\alpha=1$, then the area of detection is the quadrilateral shaded in red, yellow or blue.

The rest of the paper is organized as follows. We treat first the 
case when the shape of a signal is known in Section \ref{sec:known}. Detection of arbitrary and smooth signals are investigated respectively in Sections \ref{sec:unknown} and \ref{sec:smooth}. 
We conclude with some remarks and discussions in Section \ref{sec:disc}. All the proofs are relegated to Section \ref{sec:proof}.

\section{Detection of Signals of A Known Shape}
\label{sec:known}

We  shall assume  throughout the paper that $\sigma^2$ is known. Since the focus is on the case of short and sparse signals, when $\sigma^2$ is unknown, it can be conveniently and accurately estimated, for example, by the median absolute deviation estimator without affecting our discussions and results.  We begin with the basic notation and definitions.

We consider first the problem of detecting the signal segments, which can be cast in the framework of hypothesis testing. To fix ideas, we shall focus primarily on the case when there is a signal segment. Write $\bmu=(\mu_1,\ldots,\mu_n)^{\top}$ and denote by $\bmu_f\in \bR^n$ the mean vector specified as in (\ref{eq:signal}). More specifically,
$$
\mu_{fi}=\left\{\begin{array}{ll} f(d^{-1}(i-a))& {\rm if\ } a< i\le b\\ 0&{\rm otherwise}\end{array}\right..
$$
Let $\Delta$ be a test based on the observations $\{X_1,\ldots,X_n\}$. The null hypothesis $H_0$ is accepted when $\Delta=0$, and $H_0$ is rejected when $\Delta=1$. The probability of  the type I error  is given by
$$
\alpha(\Delta)=\bP(\Delta=1|\bmu=\zero).
$$
For a given class of signals,  the maximum probability of the type II error is represented by
$$
\beta(\Delta; \Fcal)=\sup_{f\in \Fcal}\bP(\Delta=0|\bmu=\bmu_f).
$$
We say that a test $\Delta$ is consistent for detecting signals in $\Fcal$ if
\begin{equation}
\label{eq:defdet}
\alpha(\Delta)+\beta(\Delta; \Fcal)\to 0;
\end{equation}
and signals from $\Fcal$ detectable if there exists a consistent test $\Delta$ for it. On the other hand,  a test $\Delta$ is powerless for detecting signals in $\Fcal$ if
$$
\alpha(\Delta)+\beta(\Delta; \Fcal)\to 1;
$$
and signals from $\Fcal$ is undetectable if
\begin{equation}
\label{eq:defundet}
\inf_{\tilde{\Delta}}\left\{\alpha(\tilde{\Delta})+\beta(\tilde{\Delta}; \Fcal)\right\}\to 1,
\end{equation}
where the infimum is taken over all tests based on the observations $\{X_1,\ldots,X_n\}$.

When the shape of $f$ is known in advance, then $f$ can be written as $f=Af_0$ where $f_0$ is a known function defined on $[0, 1]$ with $\int f_0^2=1$ and $A>0$ is the amplitude of $f$. Of particular interests here are the effects of the length $d$ of a signal and its  amplitude $A$ on its detectability. It is clear that signals with longer duration or larger amplitude are easier to detect. Denote by
$$
\Fcal_1(f_0, r):=\{Af_0: A\ge r\}
$$
all signals of shape $f_0$ with amplitude at least $r$ for a $r>0$,. We call  $\gamma_n(d)$  the optimal rate of detection of signals from $\Fcal_1$ with length $d$ if there exist constants $0<\underline{c}\le \overline{c}<\infty$ such that there is a test $\Delta$ that can detect any signal $f\in \Fcal_1(f_0,\bar{c}\gamma_n(d))$ with $|S|=d$
in the sense of (\ref{eq:defdet}); and yet any test is powerless for signals from $\Fcal_1^c(f_0,\underline{c}\gamma_n(d))$ with $|S|=d$ where
$$
\Fcal_1^c(f_0, r):=\{Af_0: A\le r\},
$$
in the sense of (\ref{eq:defundet}). The problem of detecting short constant signal segments, which has received much recent attention, is a special case with $f_0(x)=1$ for $x\in[0,1]$. See, e.g., Arias-Castro, Donoho and Huo (2005), Jeng, Cai and Li  (2010) and the references therein. 

As in the case of detecting a constant signal, a natural approach to the detection of a signal of a known shape is to use the  log-likelihood ratio statistics. Note that for a given interval $(j,k]$ with $0\le j<k\le n$, 
\begin{equation}
\label{eq:defknowstat}
{L}_{jk}:=\left(\sigma^2\sum_{i=1}^{k-j}f_0^2(i/d)\right)^{-1/2}\sum_{j<i\le k} X_if_0((i-j)/d),
\end{equation}
measures the log-likelihood that a signal is contained on the interval $(j,k]$, up to a scaling factor. To account for not knowing the location of a signal, we take the largest among all such likelihood ratio statistics. We note that this is commonly known as the generalized likelihood ratio test or scan statistic. Denote by ${\Delta}_n$ the detector that rejects $H_0$ if and only if ${L}_n\ge 2((1+\delta)\log n)^{1/2}$ for an arbitrary (but fixed) $\delta>0$ where
$$
{L}_n:=\max_{0\le j<k\le n} {L}_{jk},
$$
For brevity, in what follows, we shall take $\delta=0.01$. The following theorem states that the optimal rate of detection for any signal of known shape is $A \sim d^{-1/2}\log^{1/2} n$ and it is attained by the likelihood ratio test described here.

\begin{theorem}
\label{th:known}
Suppose that there is a signal of length $d<n^\xi$ for some $\xi<1$. There exists a constant $\overline{c} > 0$ for which $\Delta_n$ is consistent in testing any signal in $\Fcal_1(f_0,\overline{c}d^{-1/2}\log^{1/2} n)$. Furthermore, there exists a constant $\underline{c}>0$ for which any test is powerless in detecting signals from $\Fcal_1(f_0,\underline{c}d^{-1/2}\log^{1/2} n)$.
\end{theorem}

This theorem generalizes earlier results for the detection of constant signals. The optimal rate of detection depends on the length of the signal: the longer the signal the easier to detect.

\section{Arbitrary Signals}
\label{sec:unknown}

The aforementioned likelihood ratio tests rely heavily on the knowledge of the shape of a signal. Although appropriate in some applications where such information is available, in many other applications it may not be realistic to assume that the shape of a signal is known in advance. We now consider the detection of arbitrary signals.

In this case, it is more convenient to directly define the amplitude $A$ of a signal of length $d$ by
$$
A=\left({1\over d}\sum_{i=1}^d f^2(i/d)\right)^{1/2}.
$$
This allows us to entertain a broader class of signals that may not even be square integrable. When the signal shape is not known a priori, linear statistics similar to ${L}_{jk}$ can no longer be applied to share information across a segment. Instead, we consider the following quadratic statistic for a putative segment $(j,k]\subset \{1,\ldots, n\}$:
\begin{equation}
\label{eq:defQjk}
Q_{jk}:={1\over 2}\left[(k-j)^{1/2}+(\log n)^{1/2}\right]^{-1}\sum_{j<i\le k} \left(X_i^2/\sigma^2-1\right),\qquad \forall \; 0<j<k\le n.
\end{equation}
Again,  we take the largest among all such statistics to account for not knowing the location of a signal.
Let $T_n$ be the detector that rejects $H_0$ if and only if $Q_n\ge 2(1+\delta)\sqrt{\log n}$ where
$$Q_n:=\max_{0\le j<k\le n} Q_{jk}.$$
We now show that such a detector achieves the optimal rate of detection if the signal is entirely unknown. To this end, denote by $\Fcal_2$ the collection of all functions defined on $[0,1]$ and write
$$
\Fcal_2(r)=\{f:[0,1]\mapsto \bR| d^{-1}\sum_{i=1}^d f^2(i/d)\ge r^2\},
$$
the set of functions from $\Fcal_2$ with amplitude at least $r$; and
$$
\Fcal_2^c(r)=\{f:[0,1]\mapsto \bR| d^{-1}\sum_{i=1}^d f^2(i/d)\le r^2\},
$$
the set of functions from $\Fcal_2$ with amplitude at most $r$.
 
The fact that an arbitrary signal could be detected is itself interesting considering that the signal cannot be consistently estimated even if its location is revealed beforehand. Similar gap between detection and estimation for arbitrary signals has also be observed by Ingster and Suslina (2003) in the case when the location of the signal is known in advance.
 
\begin{theorem}
\label{th:unknown}
Suppose that there is a signal of length $d<n^\xi$ for some $\xi<1$. There exists a constant $\overline{c} > 0$ for which $\Delta_n$ is consistent in testing any signal in $\Fcal_2(\overline{c}\gamma_n(d))$ where
$$
\gamma_n(d)=\left({\log n\over d}\right)^{1/2}+\left({\log n\over d}\right)^{1/4}.
$$
Furthermore, there exists a constant $\underline{c}>0$ for which any test is powerless in detecting signals from $\Fcal_2^c(\underline{c}\gamma_n(d))$.
\end{theorem}

It is worth noting the phase transition of the optimal rate of detection of an arbitrary signal in the length of the signal segment $d$. For shorter signal segments with $d\ll \log n$, the optimal rate of detection is $A\sim d^{-1/2}\log^{1/2} n$ which is the same as if the signal shape was known. On the other hand, for longer signals such that $d\gg \log n$, the optimal rate is $A\sim {d^{-1/4}\log^{1/4} n}$. It is clear that in terms of the optimal rate of detection, we only pay a price for not knowing the shape if a signal is long in that $d\gg \log n$.

\section{Smooth Signals}
\label{sec:smooth}

We have so far considered two ``extremal" cases: the signal shape is fully known and the signal is completely arbitrary.
In some applications, though the shape of a signal may not be known, some qualitative information on the signal is available. A common example is when  a signal is known to be smooth a priori. See, e.g., Schwartzman et al. (2011). We now consider how to effectively detect short smooth signal segments.

Denote by $\Hcal_\alpha$ the $\alpha$th order H\"older space defined on $[0,1]$ for some $\alpha>0$, that is,
$$
\Hcal_\alpha(M)=\left\{f:[0,1]\mapsto \bR| |f^{(\lfloor \alpha\rfloor)}(x)-f^{(\lfloor \alpha\rfloor)}(x')|\le M|x-x'|^{\alpha-\lfloor \alpha\rfloor}\quad \forall \; x,x'\in [0,1]\right\}.
$$
Write
$$
\Fcal_3(\alpha, M,r)=\{f\in \Hcal_\alpha(M): \|f\|_{L_2}^2:=\int f^2\ge r^2\},
$$
the collection of $\alpha$-times differentiable functions whose amplitude is at least $r$; and
$$
\Fcal_3^c(\alpha, M,r)=\{f\in \Hcal_\alpha(M): \|f\|_{L_2}^2:=\int f^2\le r^2\},
$$
the collection of $\alpha$-times differentiable functions whose amplitude is at most $r$. The following result gives the lower bound for the detection boundary.

\begin{figure}[htbp]
\centerline{\includegraphics[width=0.7\textwidth,angle=270]{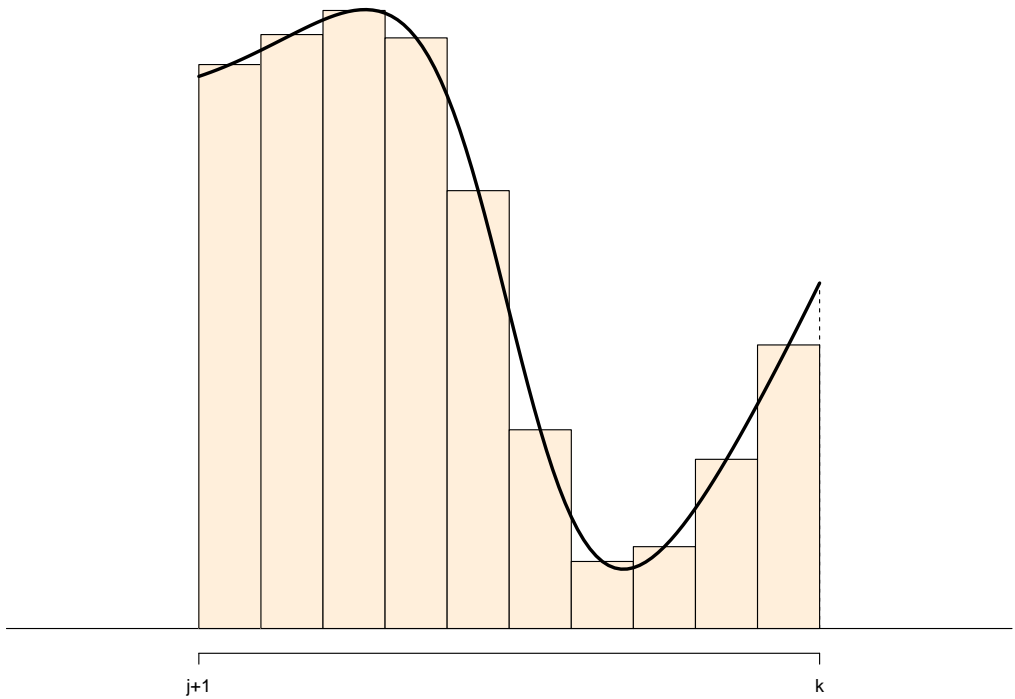}}
\vspace{-.2in}
\caption{\small Combining linear and quadratic statistics to scan for smooth signals: a segment $(j,k]$ is first divided into $l=10$ bins. With each bin, a linear statistic is computed; and all such statistics are then summarized by a quadratic statistic.}
\label{fig:linquad}
\end{figure}

\begin{theorem}
\label{th:smoothlower}
Suppose that there is a signal of length $d<n^\xi$ for some $\xi<1$. There exists a constant $\underline{c}>0$ depending on $M$ only for which any test is powerless in detecting signals from $ \Fcal_3^c(\alpha, M, \underline{c}r_n(d))$ where
$$
\gamma_n(d)=d^{-1/2}\log^{1/2} n+\min\{d^{-{2\alpha\over 4\alpha+1}}(\log n)^{\alpha\over 4\alpha+1},d^{-1/4}\log^{1/4} n\}.
$$
\end{theorem}
It is clear that when $d\ll \log n$, the optimal rate of detection remains $d^{-1/2}\log^{1/2} n$ 
and can be attained by the detector for arbitrary signals $T_n$ introduced in Section \ref{sec:unknown}. However, it turns out that $T_n$ is not a rate optimal detector of smooth signals when $d\gg \log n$ as $T_n$ does not use any information on the smoothness of the signal. To achieve the optimal rate of detection in this case, one needs to use a hybrid detector which uses both  linear and quadratic statistics. We start by considering a fixed interval $(j,k]$. The strategy is illustrated by Figure \ref{fig:linquad}.

\begin{figure}[htbp]
\centerline{\includegraphics[width=0.7\textwidth,angle=270]{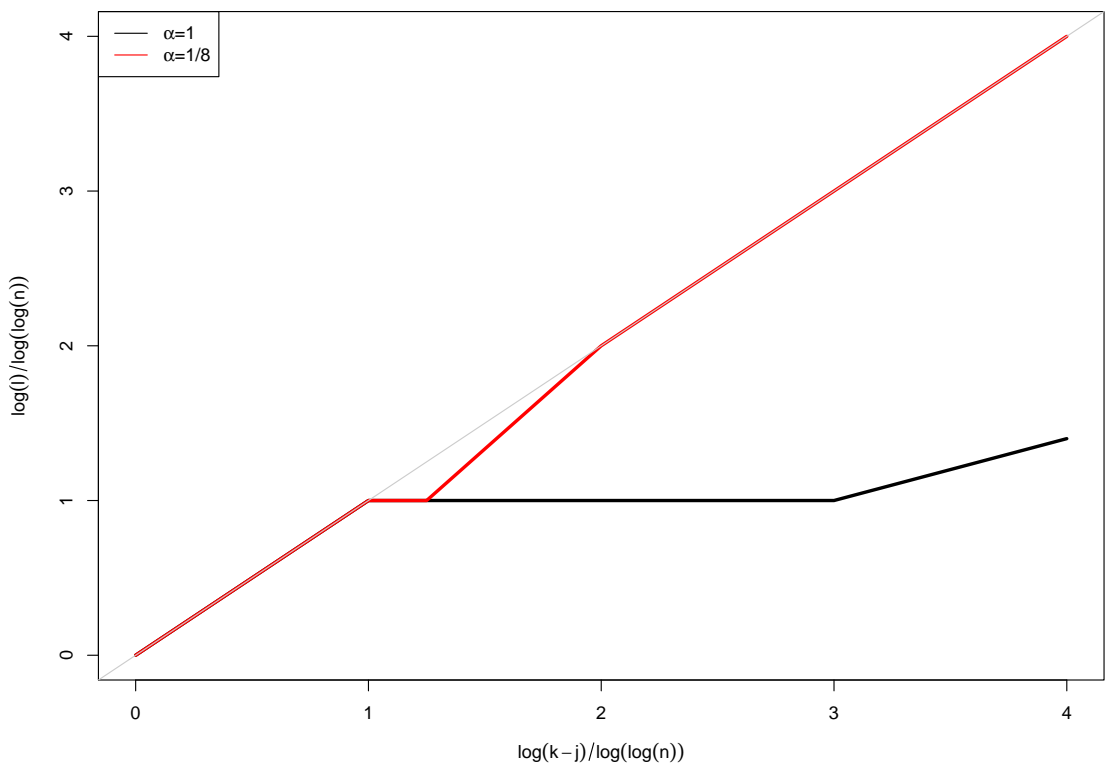}}
\vspace{-.2in}
\caption{\small Choice of the number of bins for a segment $(j,k]$ with different degrees of smoothness ($\alpha=1$ or $1/8$). Grey line corresponds to the choice of $l=k-j$ and is added for reference.}
\label{fig:choosel}
\end{figure}

To take advantage of the smoothness of a signal, we first divide the segment $(j,k]$ into bins of size $m_{jk}$ to be specified later, denoted by $B_s=(j+(s-1)m_{jk},\max\{j+sm_{jk},k\}]$ for $s=1,\ldots, l$ where $l_{jk}=\lceil (k-j)/m_{jk}\rceil$. For brevity, we shall omit the subscript of $m$ and $l$ in what follows when no confusion occurs. The intuition is that for each bin, the signal is close to a constant due to smoothness. Observe that if the signal is near constant in a segment $B_s$, the linear statistic 
$$Y_s={1\over |B_s|^{1/2}}\sum_{i\in B_s} X_i.$$
is powerful. However, across the bins, there may be considerable fluctuation and a quadratic statistic such as the one given in \eqref{eq:defQjk} is more powerful. We thus summarize the signal information on the interval $(j,k]$ by
\begin{equation}
\label{eq:defWjkm}
W_{jkl}={1\over 2}(l^{1/2}+(\log n)^{1/2})^{-1}\sum_{s=1}^l \left(Y_s^2/\sigma^2-1\right).
\end{equation}
Same as before,  we take the largest among all such statistics to account for not knowing the location of a signal. We reject $H_0$ if and only if $W_n\ge 2(1+\delta)\sqrt{\log n}$, where
\begin{equation}
\label{eq:defW}
W_n:=\max_{0\le j<k\le n} W_{jkl}.
\end{equation}
The number of bins $l$ is chosen as follows. If $\alpha\ge 1/4$, 
\begin{equation}
\label{eq:defllarge}
l=\left\{\begin{array}{ll} k-j & {\rm if\ } k-j\le \log n\\ \log n& {\rm if\ } \log n<k-j\le (\log n)^{2\alpha+1}\\ (k-j)^{2\over 4\alpha+1}(\log n)^{-{1\over 4\alpha+1}}& {\rm if\ }k-j>(\log n)^{2\alpha+1}\end{array}\right.;
\end{equation}
and if $\alpha<1/4$, we set
\begin{equation}
\label{eq:deflsmall}
l=\left\{\begin{array}{ll} k-j & {\rm if\ } k-j\le \log n\\ \log n& {\rm if\ } \log n<k-j\le (\log n)^{2\alpha+1}\\ (k-j)^{2\over 4\alpha+1}(\log n)^{-{1\over 4\alpha+1}}& {\rm if\ }(\log n)^{2\alpha+1}<k-j\le (\log n)^{1\over 1-4\alpha}\\ k-j& {\rm if\ } k-j>(\log n)^{1\over 1-4\alpha}\end{array}\right..
\end{equation}
The choice of the number of bins is illustrated in Figure \ref{fig:choosel}. 

The following theorem shows that such a detector is indeed rate optimal for signals of length $\log n\ll d< n^\xi$ for some $\xi<1$.

\begin{theorem}
\label{th:smoothupper}
Suppose that there is a signal of length $d<n^\xi$ for some $\xi<1$. There exists a constant $\overline{c}>0$ depending on $M$ only for which $W_n$ is consistent in testing any signal in $\Fcal_3(\alpha, M,\overline{c}\gamma_n(d))$ where
$$
\gamma_n(d)=d^{-1/2}\log^{1/2} n+\min\{d^{-{2\alpha\over 4\alpha+1}}(\log n)^{\alpha\over 4\alpha+1},d^{-1/4}\log^{1/4} n\}.
$$
\end{theorem}

Combining Theorems \ref{th:smoothlower} and \ref{th:smoothupper}, we can see that the optimal rate of detection for an $\alpha$ H\"older signal is
$$
A\sim \left\{\begin{array}{ll}d^{-1/2}\log^{1/2} n & {\rm if\ }d=O((\log n)^{2\alpha+1})\\ d^{-{2\alpha\over 4\alpha+1}}(\log n)^{\alpha\over 4\alpha+1}& {\rm if\ } d\gg (\log n)^{2\alpha+1}\end{array}\right.,
$$
when $\alpha\ge 1/4$; and
$$
A\sim \left\{\begin{array}{ll}d^{-1/2}\log^{1/2} n & {\rm if\ }d=O((\log n)^{2\alpha+1})\\ d^{-{2\alpha\over 4\alpha+1}}(\log n)^{\alpha\over 4\alpha+1}& {\rm if\ } d\gg (\log n)^{2\alpha+1} {\rm \ and\ } d=O((\log n)^{1/(1-4\alpha)})\\ d^{-1/4}\log^{1/4} n& {\rm if\ } d\gg (\log n)^{1/(1-4\alpha)}\end{array}\right.,
$$
when $\alpha<1/4$.

We note that for a range of segment lengths, more specifically when the length of a segment $(j,k]$ is $O\left((\log n)^{2\alpha+1}\right)$, $l=\min\{k-j,\lceil\log n\rceil\}$ is the optimal choice of the number of bins. The fact that such a choice is independent of the value of $\alpha$ offers great practical appeal since oftentimes the knowledge of $\alpha$ may be absent. For example, in many applications, there may be prior information that the length of the signal is at most $L=O((\log n)^{2\alpha_0+1})$ for some $\alpha_0>0$. Then it suffices to scan only those segments whose length is no greater than $L$, leading to the following variant of $W_n$:
$$
\tilde{W}_{nL}=\max_{\substack{0\le j<k\le n\\ k-j\le L}} W_{jkl_{jk}},
$$
where $l_{jk}=\min\{k-j,\lceil\log n\rceil\}$. As before, we claim the presence of signals and reject $H_0$ if $\tilde{W}_{nL}>2(1+\delta)\sqrt{\log n}$. It can then be shown that, not only that the computational complexity can be significantly reduced, the detector can also adaptively achieve the optimal rate of detection over all signals that are at least $\alpha_0$ times differentiable. More precisely,
\begin{theorem}
Assume that $L=O((\log n)^{2\alpha_0+1})$. Then there exists a constant $c>0$ depending on $M$ only such that for any $\alpha\ge \alpha_0$, any signal from $\Hcal_\alpha(M)$ with amplitude
$$
A\ge c\left(d^{-1/2}\log^{1/2} n+\min\{d^{-{2\alpha\over 4\alpha+1}}(\log n)^{\alpha\over 4\alpha+1},d^{-1/4}\log^{1/4} n\}\right)
$$
can be detected using $\tilde{W}_{nL}$.
\end{theorem}

\section{Discussions}
\label{sec:disc}

In this paper we considered detection of very short signal segments in three settings: signals with known shape, arbitrary signals, and smooth signals. It is of interest to note that the optimal detection rates for smooth signals connect with the cases when the signal is either of known shape or arbitrary. Smoothness diminishes when $\alpha$ decreases, and as a result, the optimal rate of detection for arbitrary signals can be viewed as the limit of that for smooth signals with $\alpha \downarrow 0$. At the other end of the spectrum, when $\alpha\uparrow \infty$, the optimal rate of detecting an $\alpha$ times differential signal becomes closer to that of detecting a signal of known shape.

To fix ideas, we have focused on the setting of Gaussian noise in the present paper. The methods can be extended to the case of  random noise with a general unknown continuous distribution by employing the binning and local median approach originally developed for nonparametric regression in Brown, Cai and Zhou (2008) and Cai and Zhou (2009), as was done in Cai, Jeng and Li (2012) for robust detection of short constant signal segments. In the current setting of general signal segments, this extension is technically much more involved than in the case of constant segments and we leave this as future work.

When the existence of a signal is detected, it is often of interest to identify the location of a signal segment. Such is the case, for example, in the CNV analysis in genomics. Intuitively, the location of detectable signals could be associated with the segments of largest scan statistics. Unlike constant signal segments, however, identification of signals of unknown shape is much more subtle because the ambiguity in defining a signal. For example, suppose that the signal segment is on the subinterval $S$, i.e., 
\begin{equation}
\label{eq:defSig}
\mu_i=\left\{\begin{array}{ll}f((i-a)/|S|)& {\rm if\ } i\in S\\ 0& {\rm otherwise}\end{array}\right.
\end{equation}
for some $S=(a,b]$. In this case, a signal $f(x)=1$ located at $(a,b]$ can also be viewed as a signal of the form $f(x)=I(0\le x\le 1/2)$ located at $(a,2b-a]$. In general, consistent estimate of the signal segment $S$ may not be as meaningful as in the case of constant signal segments because the definition of a signal itself may become ambiguous.


\section{Proofs}
\label{sec:proof}

We now prove the main results given in the paper.

\subsection{Detecting Signals of known shape}

We first prove Theorem \ref{th:known}. The argument for detecting signals of known shape is similar to those for constant signals. See, e.g., Arias-Castro, Donoho and Huo (2005).
\subsubsection{Lower bounds}

To establish the lower bounds, we consider inserting a signal $f=\gamma f_0$ to a segment of length $d$. Denote by $h_j$ the joint density of $X_1,\ldots,X_n$ when the signal is inserted to segment $((j-1)d+1,jd]$ for $j=1,\ldots, \lfloor n/d\rfloor$, and $h_0$ the joint density when there is no signal. Let $g$ be the mixture of $h_j$ for $j=1,\ldots, \lfloor n/d\rfloor$:
$$
g=\lfloor n/d\rfloor^{-1} \sum_{j=1}^{\lfloor n/d\rfloor} h_j.
$$
It can be computed that the $\chi^2$ affinity between $h_0$ and $g$ is
$$
\int \left(g^2\over h_0\right)=1+\lfloor n/d\rfloor^{-1} \left(e^{\gamma^2\sum_{i=1}^d f_0^2(i/d)}-1\right).
$$
Recall that
$$
{1\over d}\sum_{i=1}^d f_0^2(i/d)\to \int f_0^2=1
$$
as $n\to \infty$ (and consequently $d\to \infty$). This implies that $\chi^2$ affinity between $h_0$ and $g$ converges to $1$ if $h_0$ and $g$ cannot be separated if $\gamma^2\le \underline{c}d^{-1}\log n$ for sufficiently small constant $\underline{c}>0$, meaning that the sum of type I and type II error of any test converges to 1.

\subsubsection{Upper bounds}

We now show that ${\Delta}_n$ is consistent. Observe that under $H_0$, ${L}_{jk}\sim N(0,1)$. Therefore, an application of union bounds yield
$$
\bP\left\{{L}_n\ge 2((1+\delta)\log n)^{1/2}\right\}\le n^{-2\delta}.
$$
On the other hand, under the alternative $H_1$,
$${L}_{ab}\sim N\left(A\sigma^{-1}\left(\sum_{i=1}^d f_0^2(i/d)\right)^{1/2},1\right).$$
Observe that
$$
{1\over d}\sum_{i=1}^d f_0^2(i/d)\to \int f_0^2=1,\qquad {\rm as\ }d\to \infty.
$$
Therefore, for sufficiently large $n$ (and consequently $d$),
$$
{A\sigma^{-1}}\left(\sum_{i=1}^d f_0^2(i/d)\right)^{1/2}> A\sigma^{-1}\sqrt{d}/2\ge \sigma^{-1}\sqrt{\bar{c}\log n}/2.
$$
By taking constant $\bar{c}>0$ large enough, we can ensure that
$$
\bP\left\{{L}_{ab}\ge 2((1+\delta)\log n)^{1/2}\right\}\to 1.
$$
The upper bound then follows.

\subsection{Detection of arbitrary signals}

We now prove Theorem \ref{th:unknown}.
\subsubsection{Lower bounds}

We first show that any test is powerless for arbitrary signals of length $d$ and amplitude $A\le  c(d^{-1}\log n+(d^{-1}\log n)^{1/2})^{1/2}$ for some constant $c>0$. We proceed by showing that a carefully inserted signal of strength $\gamma$ may not be detected where $\gamma=c(d^{-1}\log n+(d^{-1}\log n)^{1/2})^{1/2}$. To this end, let $\phi_\mu$ be the density for a univariate normal distribution with mean $\mu$ and variance $1$. Under the null hypothesis, the joint density of $X_1,\ldots, X_n$ is simply given by
$$f(X_1,\ldots,X_n)=\prod_{i=1}^n\phi_0(X_i).$$
We now insert a random signal into the sequence. The random signal takes value $\pm \gamma$ at each of the $d$ positions on a segment $S$ leading to the following mixture:
$$
g_S(X_1,\ldots,X_n)=2^{-d}\sum_{\theta_i\in \{\pm 1\}}\left(\prod_{i\notin S}\phi_0(X_i)\prod_{i\in S}\phi_{\theta_i \gamma}(X_i)\right).
$$

Let
$$
g={1\over n-d+1}\sum_{S\in\Scal_{d}}g_{S}
$$
be the density of mixture distribution with the signal located uniformly over the collection $\Scal_d$ of length $d$ intervals. Then the $\chi^2$ affinity between $f$ and $g$ can be computed:
$$
\int \left(g^2\over f\right)={1\over (n-d+1)^2}\sum_{S_1,S_2\in\Scal_{d}}\int {g_{S_1}g_{S_2}\over f}=\bE_{S_1,S_2}\left({1\over 2}e^{\gamma^2}+{1\over 2}e^{-\gamma^2}\right)^{|S_1\cap S_2|},
$$
where $\Scal_d$ is the collection of all putative segments of length $d$, and the expectation on the rightmost hand side is taken over $S_1,S_2$ that are independently and uniformly sampled from $\Scal_{d}$. Observe that
$$
\bP(|S\cap S_2|=d-j)={2(n-d-j)\over (n-d+1)^2}
$$
for any $0\le j<d$. Therefore,
\begin{eqnarray*}
\int \left(g^2\over f\right)&=&\left(1-\sum_{j=0}^{d-1}{2(n-d-j)\over (n-d+1)^2}\right)+\sum_{j=0}^{d-1}{2(n-d-j)\over (n-d+1)^2}\left({1\over 2}e^{\gamma^2}+{1\over 2}e^{-\gamma^2}\right)^{(d-j)}\\
&\le&1+{2\over n-d+1}\sum_{j=0}^{d-1}\left({1\over 2}e^{\gamma^2}+{1\over 2}e^{-\gamma^2}\right)^{d-j}\\
&\le&1+{2d\over n-d+1}\left({1\over 2}e^{\gamma^2}+{1\over 2}e^{-\gamma^2}\right)^{d},
\end{eqnarray*}
where the last inequality follows from the fact that $e^{\gamma^2}+e^{-\gamma^2}\ge 2$.

It can be derived that
$$
\int \left(g^2\over f\right)\le 1+{2d\exp(d\gamma^2)\over n-d+1}.
$$
Therefore, taking
$$
\gamma^2=c\left(\log n\over d\right) \le {1\over 2d}\log\left(n-d+1\over d\right)
$$
for a sufficiently small constant $c>0$ yields
$$
\int \left(g^2\over f\right)\le 1+2\left(d\over n-d+1\right)^{1/2}\to 1,
$$
as $n\to \infty$. This implies that we cannot distinguish $f$ and $g$ as the $\chi^2$ affinity between them can be made arbitrarily close to $1$. In other words, any test is powerless in detecting the random signal we inserted, which has amplitude
\begin{equation}
\label{eq:proofld1}
\gamma^2=c\left(\log n\over d\right).
\end{equation}

On the other hand, observe that $e^x\le 1+x+x^2$ for any $|x|\le 1$. Therefore,
$$
{1\over 2}e^{\gamma^2}+{1\over 2}e^{-\gamma^2}\le 1+\gamma^4,
$$
provided that $\gamma^2\le 1$. As a result,
$$
\int \left(g^2\over f\right)\le 1+{2d\over n-d+1}\left(1+\gamma^4\right)^{d}\le 1+{2d\exp(d\gamma^4)\over n-d+1}.
$$
Similarly to the previous case, taking
$$
\gamma^2=\min\left\{c\left(\log n\over d\right)^{1/2},1\right\}
$$
for a sufficiently small $c>0$ yields
$$
\bE_{S_1,S_2}\left({1\over 2}e^{\gamma^2}+{1\over 2}e^{-\gamma^2}\right)^{|S_1\cap S_2|}\le 1+\left(d\over n-d+1\right)^{1/2}.
$$
which implies that any test is powerless in detecting the random signal with amplitude
\begin{equation}
\label{eq:proofld2}
\gamma^2=\min\left\{c\left(\log n\over d\right)^{1/2},1\right\}.
\end{equation}

The desired lower bound now follows immediately from Equations (\ref{eq:proofld1}) and (\ref{eq:proofld2}).

\subsubsection{Upper bound}

We now show the quadratic statistic based scan test $T_n$ indeed achieves the optimal rate and can detect any signal of length $d$ and amplitude $
A^2\ge c(d^{-1}\log n+(d^{-1}\log n)^{1/2})$ for some constant $c>0$ to be specified later. Observe that under $H_0$, 
$$T_{jk}={1\over \sigma^2}\sum_{j<i\le k}X_i^2$$ 
follows a $\chi^2_{k-j}$ distribution for any $(j,k]\subset \{1,2,\ldots,n\}$. Therefore, by the tail bound for $\chi^2$ random variables (see, e.g., Massart and Laurent, 2000),
\begin{equation}
\label{eq:chisqbd}
\bP\left(T_{jk}\ge (k-j)+2\sqrt{(k-j)x}+2x|H_0\right)\le \exp(-x).
\end{equation}
Recall that
$$
2Q_{jk}={T_{jk}-(k-j)\over (k-j)^{1/2}+(\log n)^{1/2}}.
$$
Then, for any $\delta>0$,
\begin{eqnarray*}
&&\bP\left(Q_{jk}\ge 2(1+\delta)\sqrt{\log n}|H_0\right)\\
&=&\bP\left(T_{jk}\ge (k-j)+4(1+\delta)\sqrt{(k-j)\log n}+4(1+\delta)\log n|H_0\right)\\
&\le&\bP\left(T_{jk}\ge (k-j)+4\sqrt{(k-j)(1+\delta)\log n}+4(1+\delta)\log n|H_0\right)\\
&\le&n^{-2(1+\delta)},
\end{eqnarray*}
by taking $x=2(1+\delta)\log n$ in (\ref{eq:chisqbd}). An application of union bound now yields
\begin{equation}
\label{eq:alpha}
\bP\left(Q_n\ge 2(1+\delta)\sqrt{\log n}|H_0\right)\le n^{-2\delta}\to 0.
\end{equation}

Next, consider the behavior of $T_n$ under the alternative $H_1$. Assume without loss of generality that the signal is supported on $(a,b]$. Then
\begin{eqnarray*}
T_{ab}&=&{1\over \sigma^2}\sum_{i=1}^{d}\left[f\left(i/d\right)+\epsilon_{a+i}\right]^2\\
&=&{1\over \sigma^2}\sum_{i=1}^{d}f^2\left(i/d\right)+{2\over \sigma^2}\sum_{i=1}^{d}\epsilon_{a+i}f\left(i/d\right)+{1\over \sigma^2}\sum_{i=1}^{d}\epsilon_{a+i}^2\\
&=:&B_1+B_2+B_3.
\end{eqnarray*}
Observe that
$$
B_1={dA^2\over \sigma^2}\ge c(\log n+(d\log n)^{1/2}).
$$
On the other hand, $B_2$ follows a centered normal distribution with variance $4B_1$, which implies that
$$
\bP\left(B_2\le -{1\over 4}B_1\right)\le n^{-c/128}\to 0.
$$
Moreover, $B_3$ follows a $\chi^2_d$ distribution and by $\chi^2$ tail bounds,
$$
\bP\left(B_3\le d-2\sqrt{dx}\right)\le \exp(-x).
$$
Taking $x=B_1^2/(64d)$ yields
$$
\bP\left(B_3\le d-B_1/4\right)\le n^{-c^2/64}.
$$
Thus, with probability tending to one,
$$
2Q_{ab}\ge {B_1/4\over d^{1/2}+(\log n)^{1/2}}={c(\log n)^{1/2}\over 4}\ge 4(1+\delta)(\log n)^{1/2}
$$
provided that $c\ge 16(1+\delta)$. It then follows that such a signal can be detected by $T_n$ because $Q_n\ge Q_{ab}$.

\subsection{Detection of smooth signals}
Finally, we prove Theorems \ref{th:smoothlower} and \ref{th:smoothupper}.
\subsubsection{Lower bound}
We now show that no signals from $\Fcal^c_3(\alpha, \gamma)$ of length $d$ can be detected where
$$
\gamma=c\left(d^{-1}\log n+d^{-{4\alpha\over 4\alpha+1}}(\log n)^{2\alpha\over 4\alpha+1}\right)^{1/2},
$$
where $c>0$ is a constant to be determined later. To this end, we again show that a careful inserted signal of strength $\gamma$ may not be detected. Let $\varphi$ be a positive and symmetric function such that $\varphi(u)=\tilde{\varphi}((u+1)/2)$ where
$$
\tilde{\varphi}(u)=\exp\left(-{1\over 1-u^2}\right)
$$
for $u\in (-1,1)$ and zero otherwise. Write
$$m= \lceil d^{4\alpha-1\over 4\alpha+1}(\log n)^{1\over 4\alpha+1}\rceil,$$
and $l=\lfloor d/m\rfloor$. For a binary vector $\theta=\{\pm 1\}^{l}$, write
$$
\varphi_\theta(u)=\gamma\sum_{j=1}^{l}\theta_j\varphi\left(lu-(j-1)\right).
$$
It is clear that for any $\theta$, $\varphi_\theta$ is supported on $(0,1)$, and when $\gamma\le c_1l^{-\alpha}$ for a small enough constant $c_1>0$, $\varphi_\theta\in \Fcal_3^c(\alpha, M,\gamma^2)$ (see, e.g., Tsybakov, 2008). We now insert this signal into a segment
$$
S_j=((j-1)m,(j-1+l)m]
$$
for some $j=1,2,\ldots,\lfloor (n-d)/m\rfloor$ so that
$$
\mu_i=\left\{\begin{array}{ll} \varphi_\theta((i-(j-1)m)/lm) & {\rm if\ }i\in S_j\\ 0& {\rm otherwise}\end{array}\right.
$$
Denote by $p_{\theta,j}$ the joint density function of $X_1,\ldots, X_n$ with this particular vector of means. It now suffices to show that that the null hypothesis can not be distinguished from a mixture of $p_{\theta,j}$ over all $\theta\in \{\pm 1\}^{l}$ and $j\in \{1,\ldots,\lfloor (n-d)/m\rfloor\}$:
$$
p_1:={1\over 2^{l}N}\sum_{j=1}^{N}\sum_{\theta\in \{\pm 1\}^{l}} p_{\theta,j}
$$
where $N=\lfloor (n-d)/m\rfloor$. The following lemma bounds the $\chi^2$ affinity between $p_1$ and $p_0$.

\begin{lemma}
\label{le:chisqbd}
Assume that $\log n\ll  d< n^\xi$ for some constant $\xi<1$. There exists a constant $c>0$ such that for any
\begin{equation}
\label{eq:sigstr}
\gamma^2\le c\left(\left(\log n\over d\right)+\left(\log n\over d\right)^{4\alpha\over 4\alpha+1}\right),
\end{equation}
we have
$$
\int \left({p_1^2\over p_0}\right)\le 1+2\left(l\over N-l+1\right)^{1/2},
$$
where $l$ is given by (\ref{eq:defllarge}) and (\ref{eq:deflsmall}).
\end{lemma}

The proof of Lemma  \ref{le:chisqbd} is given in the Appendix. Lemma \ref{le:chisqbd} shows that the $\chi^2$ affinity between $p_1$ and $p_0$ can be arbitrarily close to one when $n$ is large enough, and the signal strength satisfies (\ref{eq:sigstr}), which subsequently implies that
$$
\inf_{\Delta}\left\{\alpha(\Delta)+\beta(\Delta,\Fcal_3^c(\alpha,M,\gamma))\right\}\to 1.
$$
In other words, signals from $\Fcal_3^c(\alpha,M,\gamma)$ of length $d$ cannot be detected.

\subsubsection{Upper bound}

Under $H_0$, $Y_l$ follows a centered normal distribution with variance $\sigma^2$. Therefore, for any segment $(j,k]$, $\sum_{s=1}^l (Y_l^2/\sigma^2)$ follows a $\chi^2_l$ distribution. Following a similar argument as before,
$$
\bP\left(W_{jk}\ge 2(1+\delta)\sqrt{\log n}|H_0\right)\le n^{-2(1+\delta)},
$$
and by union bound,
\begin{equation}
\label{eq:alpha1}
\bP\left(W_n\ge 2(1+\delta)\sqrt{\log n}|H_0\right)\le n^{-2\delta}\to 0,
\end{equation}
for any $\delta>0$.

Now consider the case when there is a signal with amplitude
$$
A^2\ge c\left(d^{-1}\log n+d^{-{4\alpha\over 4\alpha+1}}(\log n)^{2\alpha\over 4\alpha+1}\right)
$$
for some constant $c>0$ to be specified later. For brevity, assume that the bin size $m$ is a divisor of the signal length $d$. Write
\begin{eqnarray*}
T_{ab}&=&{1\over \sigma^2}\sum_{s=1}^{l}\left({1\over \sqrt{m}}\sum_{i\in B_s} \left[f\left(i-a\over b-a\right)+\epsilon_i\right]\right)^2\\
&=&{1\over m\sigma^2}\sum_{s=1}^{l}\left(\sum_{i\in B_s} f\left(i-a\over b-a\right)\right)^2+{1\over m\sigma^2}\sum_{s=1}^{l}\left(\sum_{i\in B_s} \epsilon_i\right)^2+\\
&&\qquad\qquad+{2\over m\sigma^2}\sum_{s=1}^{l}\left(\sum_{i\in B_s} f\left(i-a\over b-a\right)\right)\left(\sum_{i\in B_s} \epsilon_i\right)\\
&=:&Q_1+Q_2+Q_3.
\end{eqnarray*}

By the smoothness of $f$, it can be shown that there exist constants $c_1,c_2>0$ such that
$$Q_1\ge d\sigma^{-2}(c_1A-c_2l^{-\alpha})^2.$$
See, e.g., Ingster (1993). Recall that
$$l\asymp d^{2\over 4\alpha+1}(\log n)^{-{1\over 4\alpha+1}}.$$
By taking the constant $c$ large enough, we can ensure that
$$
Q_1\ge c_1^2d\sigma^{-2}A^2/4\ge c_3\left(\log n+d^{1\over 4\alpha+1}(\log n)^{2\alpha\over 4\alpha+1}\right),
$$
for a sufficiently large constant $c_3>0$.

Now consider $Q_2$. Similar to before, $Q_2$ follows a $\chi^2_{l}$ distribution and therefore, again by the tail bound for $\chi^2$ random variables,
$$
\bP\left(Q_2\le l-2\sqrt{\delta l\log n}\right)\le n^{-\delta}.
$$
Note that, by taking $c>0$ large enough, we can also ensure that
$$
\sqrt{l\log n}\le {1\over 4}Q_1.
$$
Finally, note that $Q_3$ follows a normal distribution with mean zero and variance
$$
{\rm var}(Q_3)={4\over m\sigma^2}\sum_{s=1}^{l}\left(\sum_{i\in B_s} f\left(i-a\over b-a\right)\right)^2=4Q_1.
$$
By the usual tail bound for normal distribution,
$$
\bP\left(Q_3\le -Q_1/4\right)\le \exp(-Q_1/128)\to 0,
$$
as $n\to\infty$. Collecting these facts, we conclude that, with probability tending to one
\begin{eqnarray*}
W_n&\ge& W_{ab}\\
&=&{1\over 2}(l^{1/2}+(\log n)^{1/2})^{-1}(T_{ab}-l)\\
&\ge& (l^{1/2}+(\log n)^{1/2})^{-1}Q_1/4\\
&>&2(1+\delta)\sqrt{\log n},
\end{eqnarray*}
provided that $c_3$ is a large enough constant.

\section*{Appendix}
\subsection*{Proof of Lemma \ref{le:chisqbd}}
We note first that we can assume without loss of generality that $(n-d)/m$ is an integer. In the case when $(n-d)/m$ is not an integer, both $p_0$ and $p_1$ are products of the densities of $X_1,\ldots_{mN}$ and $X_{mN+1},\ldots, X_n$ where $N=\lfloor (n-d)/m\rfloor$. Because the marginal distribution of $X_{mN+1},\ldots, X_n$ remains the same under $p_0$ and $p_1$, the chi-square affinity between $p_0$ and $p_1$ is  the same as the chi-square affinity between their margins of $X_1,\ldots_{mN}$.

Denote by $\phi_{\bmu}$ the density function of a multivariate normal distribution with mean $\bmu$ and identity covariance matrix. It is clear that both $p_0$ and $p_1$ are product measures:
$$
p_0=\prod_{j=1}^{N} \phi_{\zero},
$$
and
$$
p_1={1\over N-l+1}\sum_{a=0}^{N-l}\left\{\prod_{j=a+1}^{a+l} \left({1\over 2}\phi_{c_n\bzeta}+{1\over 2}\phi_{-c_n\bzeta}\right)\prod_{j\notin (a,a+l]}\phi_{\zero}\right\}=:{1\over N-l+1}\sum_{a=0}^{N-l}g_a.
$$
where
$$
\bzeta=(\varphi(m^{-1}),\varphi(2m^{-1}),\ldots,\varphi(1))^{\top}
$$

Observe that
$$
\int {p_1^2\over p_0}=\bE_{a,a'}\int {g_ag_{a'}\over p_0},
$$
where both $a$ and $a'$ are uniformly sampled from $\{0,1,\ldots,N-l\}$. It is not hard to compute
$$
\int {g_ag_{a'}\over p_0}=\left({1\over 2}e^{\gamma^2\|\bzeta\|^2}+{1\over 2}e^{-\gamma^2\|\bzeta\|^2}\right)^{(l-|a-a'|)_+},
$$
where $(x)_+=\max\{x,0\}$. Note that
$$
\bP(|a-a'|=j)={2(N-l-j)\over (N-l+1)^2}
$$
for any $0\le j<l$. Therefore,
$$
\int {g_ag_{a'}\over p_0}=\left(1-\sum_{j=0}^{l-1}{2(N-l-j)\over (N-l+1)^2}\right)+\sum_{j=0}^{l-1}{2(N-l-j)\over (N-l+1)^2}\left({1\over 2}e^{\gamma^2\|\bzeta\|^2}+{1\over 2}e^{-\gamma^2\|\bzeta\|^2}\right)^{(l-j)}.
$$

It then follows from a similar calculation as before that
$$
\int {g_ag_{a'}\over p_0}\le 1+{2l\exp(l\gamma^2\|\bzeta\|^2)\over N-l+1}.
$$
Recall that, by the smoothness of $\varphi$,
$$
\|\bzeta\|^2\le cm\|\varphi\|_{L_2}^2,
$$
for some constant $c>0$. Therefore,
$$
\int {g_ag_{a'}\over p_0}\le 1+{2l\over N-l+1}\exp(lm\gamma^2\|\varphi\|_{L_2}^2)\le 1+2\left(l\over N-l+1\right)^{1/2}
$$
by taking
$$
\gamma^2=c\left({\log n\over d}\right)\le {\log((N-l+1)/l)\over 2lm\|\varphi\|_{L_2}^2},
$$
for a small enough constant $c>0$, where we used the fact the that $d= lm$.

On the other hand, note that there exist constants $c_1,c_2>0$ such that $e^x-1-x\le c_1x^2$ for all $|x|\le c_2$. Therefore, when
$$
\gamma^2\|\bzeta\|^2\le c_2,
$$
we have
$$
{1\over 2}e^{\gamma^2\|\bzeta\|^2}+{1\over 2}e^{-\gamma^2\|\bzeta\|^2}\le 1+c_1\gamma^4\|\bzeta\|^4\le \exp(c_1\gamma^4\|\bzeta\|^4).
$$
Therefore, for large enough $n$,
\begin{eqnarray*}
\int {g_ag_{a'}\over p_0}&\le& 1+{2l\over N-l+1}\exp(c_1m^2l\gamma^4\|\varphi\|_{L_2}^4)\\
&\le& 1+2\left(l\over N-l+1\right)^{1/2}
\end{eqnarray*}
by taking
$$
\gamma^2=cd^{-{4\alpha\over 4\alpha+1}}(\log n)^{2\alpha\over 4\alpha+1}\le \left(\log((N-l+1)/l)\over 2c_1lm^2\|\varphi\|_{L_2}^2\right)^{1/2}.
$$
for a small enough constant $c>0$. The statement of Lemma \ref{le:chisqbd} now follows.

\end{document}